\documentclass{article}
\usepackage{iclr2026_conference,times}

\usepackage{amsmath,amssymb,mathtools}
\usepackage{booktabs}
\usepackage{tabularx}
\usepackage{array}
\usepackage{placeins}
\usepackage{graphicx}
\usepackage{xcolor}
\usepackage{tikz}
\usepackage{pgfplots}
\usepackage{hyperref}
\usepackage{url}
\hypersetup{hidelinks}

\usetikzlibrary{arrows.meta,positioning,fit,calc,backgrounds}
\usepgfplotslibrary{groupplots}
\pgfplotsset{compat=1.18}

\definecolor{encblue}{RGB}{52,103,166}
\definecolor{latblue}{RGB}{210,224,242}
\definecolor{fwdgreen}{RGB}{38,128,92}
\definecolor{invred}{RGB}{183,76,72}
\definecolor{sigpurple}{RGB}{122,92,168}
\definecolor{inkgray}{RGB}{74,74,74}
\definecolor{win}{RGB}{24,128,72}
\definecolor{loss}{RGB}{184,64,64}

\colorlet{jacknote}{blue}
\colorlet{chrisnote}{green}
\colorlet{demiannote}{orange}
\colorlet{lucasnote}{purple}

\newcommand{\method}{Action-Contrastive Masked Transition Modeling}
\newcommand{\methodshort}{AC-MTM}
\newcommand{\mtmmse}{MTM-MSE}
\newcommand{\actionnce}{Action-NCE}
\newcommand{\lewm}{LeWM}
\newcommand{\sigreg}{SIGReg}
\newcommand{\nor}{NoReg}
\newcommand{\accpc}{AC-CPC}

\newcommand{\z}{\mathbf{z}}
\newcommand{\aact}{\mathbf{a}}
\newcommand{\obs}{\mathbf{o}}
\newcommand{\enc}{\mathrm{enc}_{\theta}}
\newcommand{\fpred}{\mathrm{fwd}_{\phi}}
\newcommand{\inv}{\mathrm{inv}_{\psi}}
\newcommand{\rtwo}{R^2}
\makeatletter
\newcommand{\captionhere}[2]{%
  \refstepcounter{#1}%
  \@makecaption{\csname fnum@#1\endcsname}{#2}%
}
\makeatother

\title{No Gaussian Required: Contrastive Inverse Dynamics for JEPA World Models}

\author{Jack Boylan \qquad Chris Hokamp\\
Quantexa\\
\texttt{\{jackboylan,chrishokamp\}@quantexa.com}}

\iclrfinalcopy
\begin{document}
\maketitle
\fancyhead{}

\begin{abstract}
Joint-Embedding Predictive Architectures (JEPAs) learn world models by predicting
future embeddings, but the objective admits a trivial solution of a constant
encoder, so every practical system adds an anti-collapse mechanism
\citep{lecun2022path,assran2023ijepa,bardes2022vicreg,bardes2024vjepa}. LeWorldModel (\lewm{})
prevents collapse with \sigreg{}, a regularizer that forces the latent
distribution to match an isotropic Gaussian: the representation is stabilized by
prescribing what it must look like, independently of the environment it models.
We argue that the anti-collapse pressure can instead come from the transition
data itself. \method{} (\methodshort{}) keeps \lewm{}'s forward
latent-prediction objective and adds a training-only inverse-dynamics head
trained with \actionnce{}: each latent transition must identify the action that
produced it among the other actions in the batch, a discrimination task that a
collapsed encoder provably fails. The inverse branch is discarded after
training, leaving test-time encoding, forward prediction, planning, and compute
identical to \lewm{}. On four standard pixel-control tasks under a matched
planning protocol, \methodshort{} trains stably from scratch and matches
\sigreg{} on average. On the harder multi-object OGBench Visual Scene task,
results are consistent with the prescribed geometry becoming a bottleneck:
\methodshort{} reaches $80.0{\pm}2.0\%$ success versus $58.0{\pm}2.0\%$ for
\sigreg{}, improving by 20--24 points in each training seed. A single
50-episode random-policy run gives a 52\% baseline estimate. Contrastive inverse dynamics
thus provides a distribution-free anti-collapse signal that requires no target
network, stop-gradient, pretrained encoder, or reconstruction objective, and we
characterize the action-space and observability assumptions under which it
holds. We make our code available at \url{https://github.com/jackboyla/action-contrastive-jepa}.
\end{abstract}

\section{Introduction}

A useful world model should learn from experience, predict the
consequences of actions, and support planning without reward labels or
hand-designed state. Joint-Embedding Predictive Architectures (JEPAs) offer an
appealing route: an encoder maps sensory observations to a compact latent, and a
predictor models future latents conditioned on actions
\citep{lecun2022path,zhou2025dinowm,sobal2025planning}. Because prediction
happens in representation space, the model need not reconstruct pixels. Our
experiments use images, but the objective itself only requires paired
observations and actions.

\begin{figure}[t]
\centering
\includegraphics[width=0.98\linewidth]{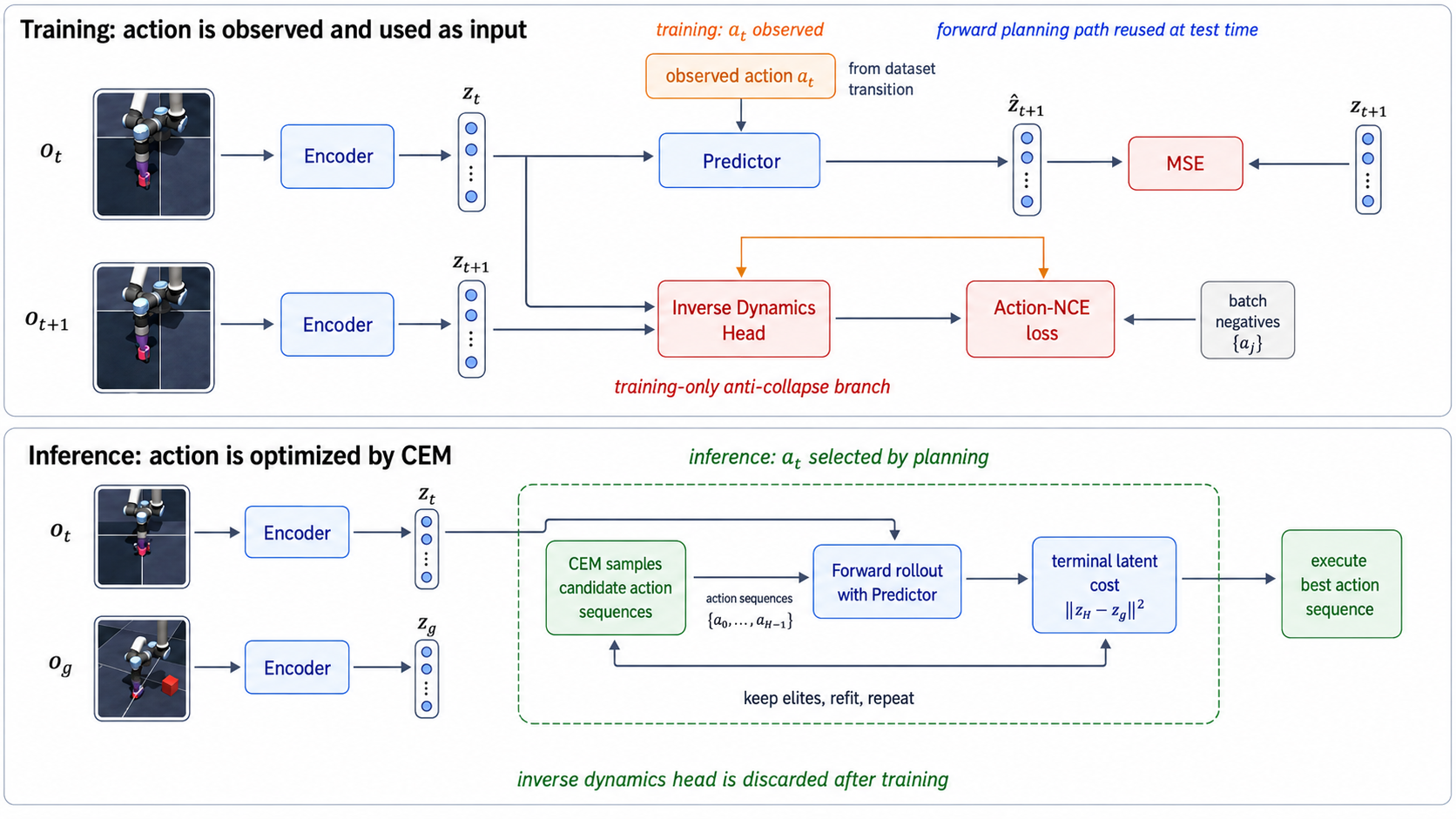}
\caption{Training and inference paths for \methodshort{}. During
training, the observed action is an input to the forward predictor and the
positive label for the contrastive inverse task. At test time, CEM samples
candidate actions and uses only the unchanged encoder and forward predictor; the
inverse dynamics/\actionnce{} branch is discarded.}
\label{fig:concept}
\end{figure}

The central difficulty is \emph{collapse}: if every observation maps to the same
latent, next-latent prediction is trivially perfect but the representation is
useless for planning. Existing systems prevent collapse by constraining the
representation with pretrained encoders, stop-gradient and
teacher--student heuristics, or explicit distributional regularizers
\citep{assran2023ijepa,bardes2022vicreg,bardes2024vjepa}. \lewm{}
\citep{maes2026leworldmodel} distills this line to its simplest form: an
end-to-end pixel JEPA trained with next-embedding prediction plus \sigreg{}, a
regularizer from LeJEPA \citep{balestriero2025lejepa} that projects embeddings
onto random directions and pushes every one-dimensional marginal toward a
Gaussian. This reduces the multi-term PLDM recipe \citep{sobal2025planning} to a
single effective coefficient.

\sigreg{} is principled, but it stabilizes the world model by prescribing a
\emph{global, isotropic-Gaussian} latent geometry that the encoder must satisfy
regardless of the environment being modeled. \citet{maes2026leworldmodel}
themselves flag this prescription as a possible cause of weak \lewm{}
performance in low-intrinsic-dimensionality environments. This motivates our
question: \emph{rather than imposing a distribution the dynamics never asked
for, can anti-collapse be derived from the transition data itself?}

Sensorimotor World Models (SMWM) answers this question with a standalone
inverse-action MSE regularizer in the same LeWM setting
\citep{ivashkov2026smwm}. Our focus is
the contrastive formulation: a chance-level collapse bound, its reliability
relative to inverse MSE, one loss weight held fixed across tasks, and a harder
Scene stress test.

We study \method{} (\methodshort{}). As in \lewm{}, the model predicts the
future latent from the current latent and action. During training only, an
inverse head additionally maps each adjacent latent pair to an action query that
must identify the observed action among the other actions in the batch, an
InfoNCE-style objective \citep{gutmann2010nce,oord2018cpc} we call \actionnce{}.
A constant encoder maps every transition to the same query, so action
identification cannot beat chance (Section~\ref{sec:method}): collapse turns
the forward objective's global optimum into a representation that cannot
outperform chance on the inverse task. The resulting
anti-collapse pressure is dynamics-native; the encoder is forced to keep
exactly the information that distinguishes the effects of actions. The
entire inverse branch is discarded after training, leaving the deployed model
identical to \lewm{}. We evaluate \methodshort{} as a controlled \lewm{}
modification against \sigreg{}, a non-contrastive inverse-regression ablation
(\mtmmse{}), and a contrastive forward-prediction alternative (\accpc{}). Three
findings emerge:

\textbf{(1) Dynamics-derived anti-collapse works, with no distributional prior
and no test-time change.} Trained from scratch under \lewm{}'s one-stage recipe,
\methodshort{} never collapses across tasks and seeds and matches \sigreg{} on
the standard four-task suite (TwoRoom, Reacher, PushT, OGBench-Cube), winning
two tasks, tying one, and losing PushT. The planner never calls the inverse
head: evaluation uses the same encoder, autoregressive predictor, latent
distance, and CEM planner as \lewm{}.

\textbf{(2) On the harder Scene task, results are consistent with the prescribed
geometry becoming a bottleneck.} On OGBench Visual Scene, where a single arm controls a drawer,
window, buttons, and a movable cube, \sigreg{} drops to a three-seed mean of
58.0\% under the matched trajectory-goal MPC protocol, while \methodshort{}
remains at 80.0\%, with gains of 24, 20, and 22 points across the three training
seeds. A single 50-episode random-policy run gives a 52\% baseline estimate.
The inverse-regression ablation also
clears \sigreg{} decisively (75.3\%), so the advantage belongs to the
dynamics-derived signal itself, not to one loss form. Both models avoid
collapse; the result may combine more useful transition geometry with more
accurate short-horizon latent dynamics. Appendix~\ref{sec:appendix-scene-audit}
audits this result.

\textbf{(3) The contrastive form of the inverse signal is what makes it
reliable.} Plain inverse regression (\mtmmse{}) is strong on TwoRoom, Cube, and
Scene, but collapses on two of three Reacher seeds; \actionnce{} removes this
bimodality at the cost of 3.8 points on a long-horizon stress test. Both inverse
variants trail \sigreg{} on PushT, where identifying the action underweights
weakly controlled object state, a limitation we analyze in Section \ref{sec:mechanisms}.

Our contributions are: (i) \methodshort{}, a distribution-free, dynamics-native
anti-collapse mechanism for end-to-end JEPA world models, with a simple
chance-level lower bound showing why collapse cannot pay; (ii) a controlled
three-seed comparison on five pixel-control tasks, matched down to the planner,
isolating the training signal as the only moving part; and (iii) probe,
stress-test, and counterfactual-surprise analyses that separate ``does not
collapse'' from ``supports planning,'' including an explicit account of when
inverse-dynamics anti-collapse should and should not be expected to hold.

\section{Background and Positioning}
\label{sec:positioning}

\paragraph{Reward-free latent planning.}
We consider offline trajectories of pixel observations and actions
$\tau=(\obs_0,\aact_0,\obs_1,\dots)$, with no rewards or optimality assumptions.
During training, $\aact_t$ is the recorded continuous command carrying
$\obs_t$ to $\obs_{t+1}$. At test time, no action label is available: a
model-predictive-control solver based on the cross-entropy method (CEM) samples,
scores, and refits a distribution over candidate action sequences to minimize
latent distance to an encoded goal
\citep{sobal2025planning,maes2026leworldmodel}.

\paragraph{From PLDM to \lewm{}.}
PLDM prevents collapse with a VICReg-derived objective of several interacting
terms \citep{sobal2025planning,bardes2022vicreg}. \lewm{} replaces it with
prediction plus \sigreg{}, matching all one-dimensional latent marginals to a
Gaussian (motivated by the Cram\'er--Wold theorem)
\citep{balestriero2025lejepa,maes2026leworldmodel}. This simplification is the
right comparison point for our work: the contribution of \lewm{} is not only
performance, but the fact that stable end-to-end pixel JEPA world models can be
made dramatically simpler than earlier recipes. We keep that agenda while
replacing the remaining global distribution-matching term with a local,
contrastive transition signal.

\paragraph{A family of distribution-free signals.}
\mtmmse{} is our SMWM-style inverse-regression baseline
\citep{ivashkov2026smwm}: it regresses the observed action from an adjacent
latent pair. SMWM establishes this basic inverse-MSE mechanism and tunes its
loss weight per environment; our controlled baseline uses one coefficient
chosen on Reacher and held fixed across tasks. To separate ``does not collapse'' from ``plans well,'' we
also evaluate \accpc{}, an action-conditioned contrastive \emph{forward}
objective that identifies the true future latent among batch negatives rather
than identifying the action. All variants are training signals plugged into the
same \lewm{} encoder, predictor, and planner; Table~\ref{tab:simplicity}
summarizes what each signal asks of the representation.

\begin{table}[t]
\centering
\small
\begin{tabularx}{\textwidth}{lXl}
\toprule
Method & Anti-collapse signal & Latent geometry prescribed? \\
\midrule
PLDM \citep{sobal2025planning} & VICReg variance/covariance (multi-term) & Yes (variance floor) \\
\lewm{} \citep{maes2026leworldmodel} & \sigreg{} isotropic-Gaussian matching & Yes (global Gaussian) \\
\mtmmse{} \citep{ivashkov2026smwm} & Inverse-action MSE $(\z_t,\z_{t+1})\!\to\!\aact_t$ & No \\
\methodshort{} (ours) & Contrastive inverse-action identification & No \\
\accpc{} & Contrastive future identification & Implicit (unit sphere) \\
\bottomrule
\end{tabularx}
\caption{Anti-collapse mechanisms compared in this paper. \methodshort{} uses
negatives during training, but does not prescribe a global latent marginal and
does not change the test-time planner.}
\label{tab:simplicity}
\end{table}

\section{Action-Contrastive Masked Transition Modeling}
\label{sec:method}

Let latent embedding of observation $\obs_t$ at timestep $t$ be $\z_t=\enc(\obs_t)$. We write a one-step predictor for compactness; the
implementation inherits \lewm{}'s causal latent-history predictor. The forward (planning) task is
\begin{equation}
  \hat{\z}_{t+1} = \fpred(\z_t,\aact_t), \qquad
  \mathcal{L}_{\mathrm{fwd}} = \big\|\hat{\z}_{t+1}-\z_{t+1}\big\|_2^2 .
\end{equation}
\methodshort{} adds a training-only inverse-dynamics head. For each of the $N$
transitions in a flattened batch/window, it produces an action query
\begin{equation}
  \hat{\aact}_i=\inv(\z_i,\z_{i+1}).
\end{equation}
We score every observed action $\aact_j$ as a candidate for transition $i$ using
negative squared distance, giving an InfoNCE-style classification loss
\citep{gutmann2010nce,oord2018cpc} over in-batch actions:
\begin{equation}
  s_{ij}=-\frac{\|\hat{\aact}_i-\aact_j\|_2^2}{\tau d_a},
  \qquad
  \mathcal{L}_{\mathrm{NCE}}
  =-\frac{1}{N}\sum_{i=1}^N
  \log\frac{\exp(s_{ii})}{\sum_{j=1}^N\exp(s_{ij})},
  \label{eq:action-nce}
\end{equation}
where $d_a$ is the action dimension. The total objective is
\begin{equation}
  \mathcal{L}_{\methodshort}
  =\mathcal{L}_{\mathrm{fwd}}+\lambda\mathcal{L}_{\mathrm{NCE}},
  \qquad \lambda=0.30,\quad \tau=0.10.
  \label{eq:total}
\end{equation}
There is no \sigreg{} term. The fixed coefficient was selected by a bounded
Reacher stability sweep and then used unchanged across all tasks.

``Masked'' refers to factor prediction within a transition tuple, not image
patch masking. The forward task withholds $\z_{t+1}$ and predicts it from
$(\z_t,\aact_t)$; the inverse task withholds $\aact_t$ and predicts it from
$(\z_t,\z_{t+1})$. We optimize both tasks on every batch rather than sample one
mask at a time.

The candidate set in Equation~\ref{eq:action-nce} is the $N{=}B(T{-}1)$ observed
action blocks already present in the batch; any dataset action or a memory bank
could serve instead, but we use in-batch actions for several reasons. First,
the blocks are already resident on the GPU, so the negatives add only an
$N\times N$ distance matrix. Second, the candidates $\aact_j$ are \emph{raw}
actions, not encoder outputs, so they carry no gradient to $\enc$; enlarging the
pool changes only the difficulty of the discrimination, not the gradient path.
Third, the objective is anti-collapse, which the in-batch pool already enforces
through the $\log N$ floor of Equation~\ref{eq:nce-floor}. Sampling globally would
mostly add easy negatives (blocks far from the prediction that contribute
negligible gradient) while raising the rate of false negatives, since
control actions repeat (near-zero or saturated blocks) and a duplicated
``negative'' penalizes a correct prediction. A larger candidate set therefore adds
compute and label noise to sharpen an action-retrieval property the method does
not require.

\paragraph{Why \actionnce{} opposes collapse.}
Forward prediction alone cannot distinguish a useful representation from a
constant one: if $\enc(\obs)=c$ for every observation and
$\fpred(c,\aact)=c$, then $\mathcal{L}_{\mathrm{fwd}}=0$. \actionnce{} turns this
degenerate solution into a failed classification problem. Under collapse, every
transition gives the inverse head the same input pair $(c,c)$, so every row of
the action classifier is identical. The model is then forced to assign one fixed
probability vector $p$ to all $N$ positives in the batch. Since each candidate
action is the correct label exactly once, the average loss satisfies
\begin{equation}
  -\frac{1}{N}\sum_{i=1}^{N}\log p_i \ge \log N,
  \label{eq:nce-floor}
\end{equation}
with equality only at the chance classifier. Thus collapse can drive the forward
loss to zero but cannot drive the contrastive inverse loss below chance. To
improve \actionnce{}, the encoder must preserve transition information that makes
the observed action more identifiable than the in-batch alternatives.

\paragraph{Why not use non-contrastive inverse regression?}
\mtmmse{} is the SMWM-style distribution-free replacement for \sigreg{}
\citep{ivashkov2026smwm}: it replaces
Equation~\ref{eq:action-nce} with
$\mathcal{L}_{\mathrm{inv}}=\|\hat{\aact}_t-\aact_t\|_2^2$. It is an important
ablation because it asks whether any inverse-dynamics signal is enough.
\mtmmse{} is strong on TwoRoom, PushT, and Cube, but the instability appears on
Reacher, where two of three seeds collapse. The reason is that inverse
regression supplies only a variance-scale floor under complete collapse: its
best collapsed prediction is the mean action, and the loss is the action
variance. When multiple action blocks can produce similar visual endpoints, that
margin can be weak, while forward MSE still rewards shrinking the latent scale.
\actionnce{} keeps the same training-only inverse head but changes the failure
geometry to chance-level identification, which gives a sharper barrier against
constant latents. Section~\ref{sec:stability} analyzes this failure mode.

\paragraph{Controlled implementation.}
We inherit the \lewm{} vision encoder, latent projector, causal forward
predictor, action conditioning, offline datasets, and CEM planner. The only added
component is a small MLP inverse head used during training. After training, the
head is never called by model rollout or cost evaluation and can be removed
without changing predictions. Consequently \methodshort{} and \lewm{} use the
same test-time computation and planner.

\section{Experiments}
\label{sec:experiments}

\begin{figure}[t]
\centering
\begin{tabular}{@{}ccccc@{}}
\includegraphics[width=0.17\linewidth]{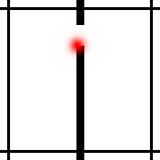} &
\includegraphics[width=0.17\linewidth]{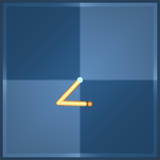} &
\includegraphics[width=0.17\linewidth]{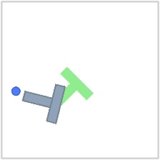} &
\includegraphics[width=0.17\linewidth]{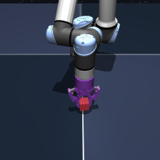} &
\includegraphics[width=0.17\linewidth]{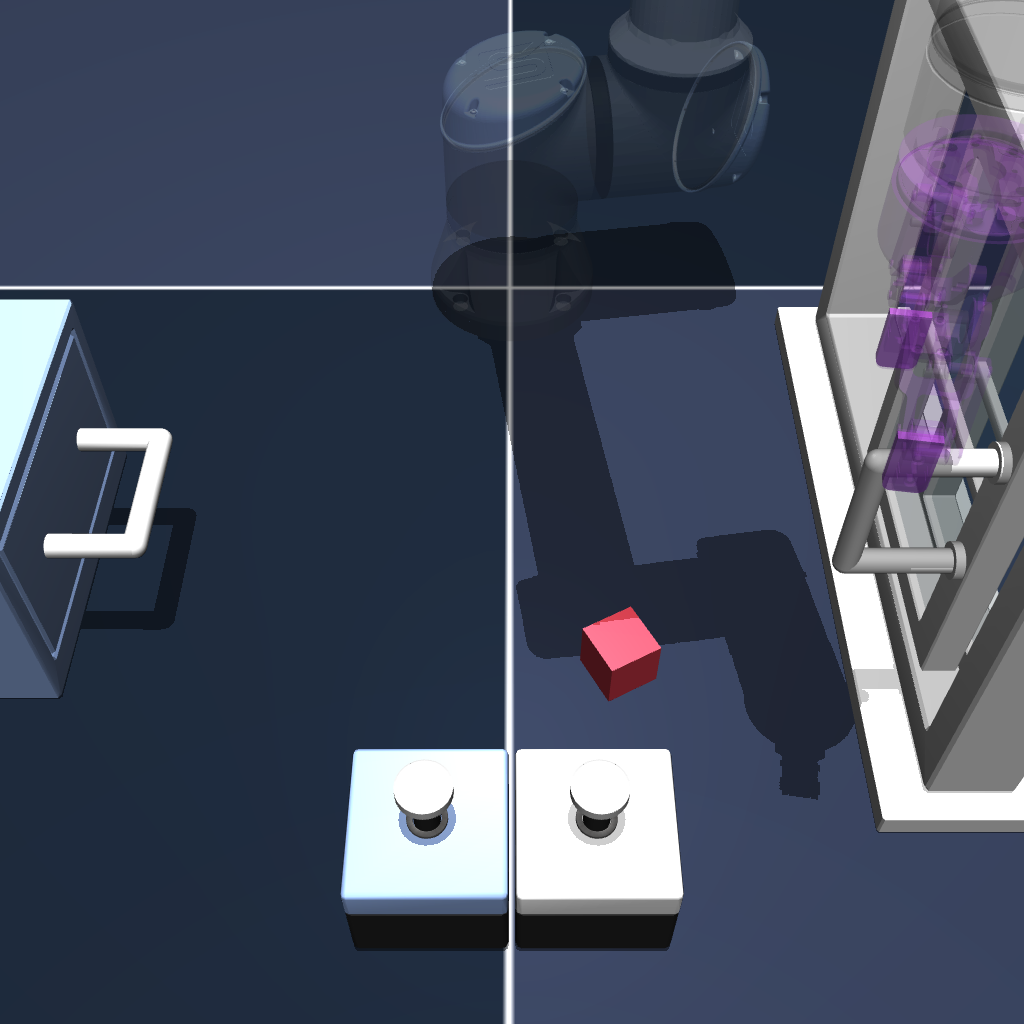} \\
\textbf{TwoRoom} & \textbf{Reacher} & \textbf{PushT} & \textbf{OGBench-Cube} &
\textbf{OGBench-Scene}
\end{tabular}
\caption{Evaluation suite. TwoRoom is low-dimensional
navigation through a wall opening; Reacher matches a target arm configuration;
PushT requires contact-rich pushing of a T-shaped block to a goal pose;
OGBench-Cube is 3D robot-arm cube manipulation; OGBench-Scene adds a multi-object
drawer/window/button/cube scene. These tasks stress different failure modes: low
intrinsic dimension, articulated dynamics, contact, 3D manipulation, and
multi-object visual complexity.}
\label{fig:tasks}
\end{figure}

\paragraph{Evaluation Setup.}
We evaluate the \lewm{} continuous-control suite. This includes TwoRoom, Reacher,
PushT, and OGBench-Cube. We add OGBench Visual Scene as a harder multi-object
manipulation study (Figure~\ref{fig:tasks}). All models train
end-to-end from pixels in one stage. We evaluate each policy with the same
CEM/MPC planner used by \citet{maes2026leworldmodel}. CEM samples 300 candidate
action sequences, retains 30 elites, and refits the sampling distribution for 30
iterations. In Figure~\ref{fig:multitask}
we report \methodshort{} and \mtmmse{} against our matched \sigreg{}
reproduction of \lewm{}, alongside paper-reported PLDM, DINO-WM,
goal-conditioned policy, and random baselines from
\citet{maes2026leworldmodel}; on OGBench-Scene, where no external numbers
exist, we evaluate the random policy ourselves under the identical protocol.
Our controlled claims use three training seeds
$\{3072,1,2\}$, 200 evaluation episodes, evaluation seed 42,
goal offset 25, and interaction budget 50. TwoRoom-long changes only the goal offset and
budget to $100/150$. The OGBench-Scene task uses the same trajectory-goal MPC protocol
as the other visual tasks, with 50 evaluation episodes per training seed.
Appendix~\ref{sec:appendix-protocol} records the
full protocol.

\subsection{Collapse sanity check}
On TwoRoom, removing \sigreg{} without replacement (\nor{}) drives forward loss to
$\approx 0$ in training diagnostics, the trivial constant-latent solution
described above. Both \mtmmse{} and \methodshort{} inverse tasks prevent this failure while preserving the
same one-stage training recipe proposed by \citet{maes2026leworldmodel}. This confirms that transition supervision, not
merely removal of \sigreg{}, provides the anti-collapse signal
(Table~\ref{tab:noreg}).

\begin{table}[t]
\centering
\small
\begin{tabular}{lc}
\toprule
Model & TwoRoom success (\%) \\
\midrule
\nor{}: no anti-collapse term & $28.0{\pm}2.0$ \\
\lewm{} with \sigreg{} & $85.5{\pm}0.4$ \\
\mtmmse{} & $90.2{\pm}0.5$ \\
\methodshort{} & $\mathbf{90.7{\pm}0.6}$ \\
\bottomrule
\end{tabular}
\caption{TwoRoom anti-collapse ablation with 200 evaluation episodes
per seed. Values are mean $\pm$ standard deviation over three training seeds.
Plain next-latent prediction reaches a trivial low-loss solution and plans
substantially worse.}
\label{tab:noreg}
\end{table}

\subsection{Planning results in the LeWM evaluation frame}
\label{sec:multitask}
\begin{figure}[t]
\centering
\includegraphics[width=\linewidth]{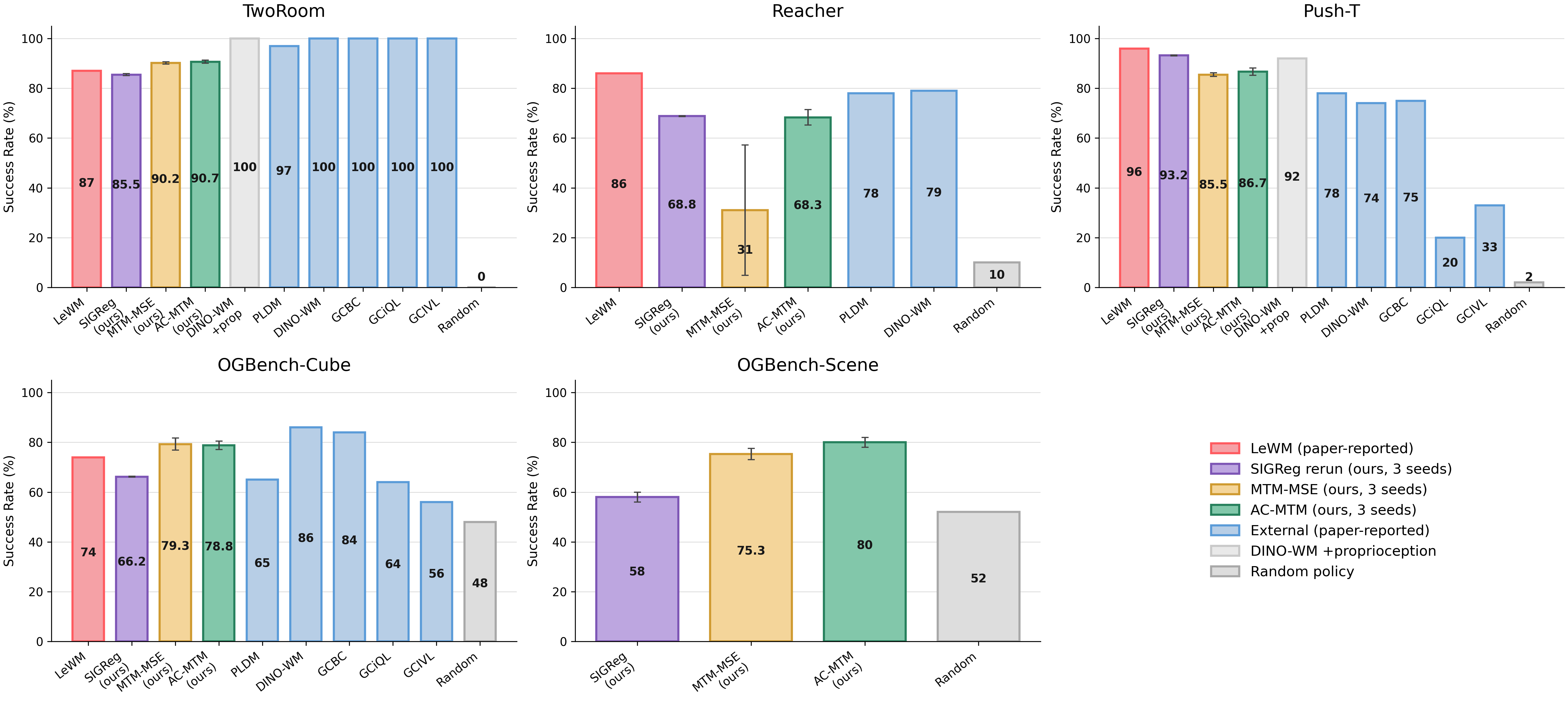}
\caption{Planning success across the five environments. Salmon and blue
bars are paper-reported LeWM/external baselines from
\citet{maes2026leworldmodel}; purple, gold, and green bars are our controlled
\sigreg{}, \mtmmse{}, and \methodshort{} evaluations across three training
seeds. No external baseline has published OGBench-Scene results under this
protocol, so the Scene panel contains only our controlled models and the random
policy. The external baselines place \methodshort{} in the original evaluation
frame; direct statistical claims are made only for the controlled comparisons in
Table~\ref{tab:results}.}
\label{fig:multitask}
\end{figure}

Table~\ref{tab:results} gives the controlled \methodshort{} versus \sigreg{}
comparison, while Figure~\ref{fig:multitask} restores the original evaluation
style by including the external paper-reported baselines. The figure is
contextual: PLDM, DINO-WM, GCBC, GCiQL, and GCIVL are taken from the \lewm{}
paper, while the statistical claims come from our matched \sigreg{} reruns.
\methodshort{} uses the same test-time planner as \lewm{} and improves over
\sigreg{} on TwoRoom, Cube, and Scene, matches it on Reacher, and trails it on
PushT. The PushT deficit is analyzed in Section~\ref{sec:mechanisms}; the full
\mtmmse{} ablation is reported in Appendix~\ref{sec:appendix-controlled}.

\begin{table}[t]
\centering
\small
\begin{tabular}{lccc}
\toprule
Task & \sigreg{} (\lewm{}) & \methodshort{} (ours) & $\Delta$ \\
\midrule
TwoRoom  & $85.5{\pm}0.4$ & $\mathbf{90.7{\pm}0.6}$ & \textcolor{win}{$+5.2$} \\
Reacher  & $68.8{\pm}0.2$ & $68.3{\pm}3.1$ & $-0.5$ \\
PushT    & $\mathbf{93.2{\pm}0.2}$ & $86.7{\pm}1.5$ & \textcolor{loss}{$-6.5$} \\
OGB-Cube & $66.2{\pm}0.2$ & $\mathbf{78.8{\pm}1.7}$ & \textcolor{win}{$+12.6$} \\
OGB-Scene & $58.0{\pm}2.0$ & $\mathbf{80.0{\pm}2.0}$ & \textcolor{win}{$+22.0$} \\
\bottomrule
\end{tabular}
\caption{Planning success (\%) with the shared autoregressive planner: 200
evaluation episodes per seed on the standard tasks and 50 on OGBench-Scene.
Values are mean $\pm$ standard deviation over three training seeds; bold marks
differences larger than the cross-seed noise. Figure~\ref{fig:multitask} places
these numbers next to external paper-reported baselines.}
\label{tab:results}
\end{table}

\subsection{Harder OGBench Scene separates the methods}
\label{sec:scene}
The largest margin appears on the more complex OGBench Visual Scene task
(Table~\ref{tab:scene}). Scene is qualitatively different from OGBench-Cube: the
same arm must control a multi-object visual state involving a drawer, window,
buttons, and movable object, so a useful latent must preserve several
slow-changing task variables at once. Under the trajectory-goal MPC protocol,
\sigreg{} falls to a three-seed mean of 58.0\% success while \methodshort{}
achieves 80.0\%. The random-policy run helps calibrate these numbers: like
OGBench-Cube (48\% random), the 25-step trajectory-goal protocol leaves many
episodes near-solved at reset. One 50-episode random-policy run scores 52.0\%
(Wilson 95\% interval, 38.5 to 65.2\%), so we treat it as a baseline estimate. The paired comparison pooled over the three repeated sets of 50
start--goal tasks gives 40 \methodshort{}-only successes versus 7
\sigreg{}-only successes. Notably, \sigreg{} does not collapse on Scene (its final latent
scale is healthy; Appendix~\ref{sec:appendix-scene-audit}); the gap is between
avoiding collapse and learning a geometry the planner can exploit.

The \mtmmse{} ablation localizes the credit. Trained on Scene under the
identical protocol, it reaches $75.3{\pm}2.3\%$ with no collapsed seed and
beats \sigreg{} in the paired comparison (39 wins vs.\ 13 losses,
$p\!\approx\!4{\times}10^{-4}$ in an episode-level descriptive test). The Scene advantage therefore belongs to the
inverse-dynamics \emph{family}: deriving anti-collapse from the transitions,
in either form, beats prescribing the latent distribution once several
controllable factors must coexist in the latent. The contrastive form adds
4.7 points over inverse regression on Scene (16 wins vs.\ 9 losses,
$p\!\approx\!0.23$ in the same descriptive test) on top of the Reacher reliability
that motivates it (Section~\ref{sec:stability}).

\begin{table}[t]
\centering
\small
\begin{tabular}{lccccc}
\toprule
Method & Seed 3072 & Seed 1 & Seed 2 & Mean & Paired wins/losses vs.\ \sigreg{} \\
\midrule
Random & -- & -- & -- & $52.0$ & -- \\
\sigreg{} & 56.0 & 58.0 & 60.0 & $58.0{\pm}2.0$ & -- \\
\mtmmse{} & 78.0 & 74.0 & 74.0 & $75.3{\pm}2.3$ &
\textcolor{win}{39 / 13} \\
\methodshort{} & 80.0 & 78.0 & 82.0 & $\mathbf{80.0{\pm}2.0}$ &
\textcolor{win}{40 / 7} \\
\bottomrule
\end{tabular}
\caption{OGBench Visual Scene trajectory-goal success (\%), with 50 evaluation
episodes per training seed and evaluation seed 42. The random policy is a
single 50-episode evaluation under the identical protocol and is a
baseline estimate. Paired wins/losses pool all 150 outcomes against \sigreg{}:
two-sided binomial/McNemar
$p\!\approx\!1.1{\times}10^{-6}$ for \methodshort{} (40/7) and
$p\!\approx\!4.1{\times}10^{-4}$ for \mtmmse{} (39/13); the
\methodshort{}--\mtmmse{} pairing is 16/9 ($p\!\approx\!0.23$). Because the
same 50 tasks recur across the three training seeds, these are episode-level
descriptive tests, not tests with 150 independent experimental units. The
seed-level gains for \methodshort{} over \sigreg{} are 24, 20, and 22 points.
This is not a
public OGBench leaderboard claim; the official fixed-goal protocol is separate.
Appendix~\ref{sec:appendix-scene-audit} audits possible explanations for the
large margin.}
\label{tab:scene}
\end{table}

\subsection{Reliability and stress-test trade-offs}
\label{sec:stability}
The Reacher seed breakdown makes the stability difference explicit
(Table~\ref{tab:reacher-seeds}). One \mtmmse{} run is competitive, but two enter
the constant-latent state. In contrast, all \methodshort{} runs remain useful.
On TwoRoom-long the non-contrastive inverse-regression ablation is stronger:
$28.0{\pm}0.8\%$ versus $24.2{\pm}0.6\%$ for \methodshort{}
(Table~\ref{tab:long}). Both \mtmmse{} and \actionnce{} remain above \sigreg{} on this
long-goal stress test, but \actionnce{} trades 3.8 points of the pure-MSE
long-horizon gain for much better Reacher reliability.

\begin{table}[ht]
\centering
\small
\begin{tabular}{lrrrr}
\toprule
Method & Seed 3072 & Seed 1 & Seed 2 & Mean \\
\midrule
\sigreg{} & 69.0 & 69.0 & 68.5 & 68.8 \\
\mtmmse{} & 68.0 & 11.5 & 13.5 & 31.0 \\
\methodshort{} & 70.5 & 70.5 & 64.0 & 68.3 \\
\bottomrule
\end{tabular}
\caption{Reacher success (\%) by training seed, with 200 evaluation episodes per
checkpoint. \mtmmse{} succeeds once but collapses twice; \methodshort{} removes
that bimodality.}
\label{tab:reacher-seeds}
\end{table}

\begin{table}[ht]
\centering
\small
\begin{tabular}{lccc}
\toprule
TwoRoom-long ($100/150$) & \sigreg{} & \mtmmse{} & \methodshort{} \\
\midrule
Success (\%) & $17.0{\pm}0.4$ & $\mathbf{28.0{\pm}0.8}$ & $24.2{\pm}0.6$ \\
\bottomrule
\end{tabular}
\caption{Long-horizon stress test with 200 evaluation episodes and three
training seeds per method. \methodshort{} retains most of the inverse-dynamics
gain over \sigreg{} but trails \mtmmse{} by 3.8 points.}
\label{tab:long}
\end{table}

\subsection{PushT probes diagnose the remaining family-level gap}
\label{sec:mechanisms}
Linear probes freeze the encoder, embed 4000 frames, and fit ridge regression
($\alpha{=}1$) from the latent to each privileged simulator-state coordinate.
Both inverse variants trail \sigreg{} on PushT, and the probes show the same
family-level failure mode for \mtmmse{} and \methodshort{}
(Table~\ref{tab:mechanism}): both preserve agent position, retain much of block
position, and underencode the T-block orientation, a weakly controlled variable
that determines contact geometry.

\begin{table}[t]
\centering
\small
\begin{tabular}{llccc}
\toprule
Mechanism & Planner & Orient.\ $\rtwo{}$ & Mean $\rtwo{}$ & Success (\%) \\
\midrule
\sigreg{} (\lewm{}) & AR & \textbf{0.791} & \textbf{0.701} & \textbf{93.2} \\
\mtmmse{} & AR & 0.508 & 0.674 & 85.5 \\
\methodshort{} & AR & 0.514 & 0.675 & 86.7 \\
\accpc{} & AR & 0.655 & 0.564 & 62.5 \\
\bottomrule
\end{tabular}
\caption{PushT anti-collapse mechanisms: frozen-latent probes ($n{=}4000$,
$\alpha{=}1$; orientation is $\mathrm{state}[4]$) and 200-episode planning.
\methodshort{} probe values are means over three training seeds; all rows use
the same autoregressive planner.}
\label{tab:mechanism}
\end{table}

\section{Analysis}
\label{sec:analysis}

\paragraph{Why \actionnce{} stabilizes Reacher.}
Inverse-MSE only resists collapse to the extent that the observed action can be
regressed below the mean-action baseline. Reacher applies a multi-step action
block to a visually observed two-link arm; many action sequences can produce
similar endpoints, so the regression margin can be weak. At the same time,
forward latent MSE rewards shrinking the online representation. \actionnce{}
changes the failure geometry: even when absolute action regression is
ambiguous, identical transition queries cannot assign different positives to
different rows and remain at chance by Equation~\ref{eq:nce-floor}. This does
not make the inverse problem fully identifiable, but it provides a stronger
early barrier against the constant-latent attractor.

\paragraph{What \actionnce{} does not solve.}
The contrastive loss still emphasizes state that separates actions. It does not
unconditionally preserve every task-relevant variable, unlike \sigreg{}'s global
variance pressure. PushT requires accurate block pose and orientation even when
the same commanded action has little visible effect before contact. This is a
plausible reason both inverse variants remain below \sigreg{}. The remaining
problem is representation coverage, not test-time action selection: every
method uses the same CEM planner.

\paragraph{Why probes are diagnostic, not decisive.}
Linear probes measure whether a quantity is linearly decodable from the
latent; planning measures whether the forward model and latent metric
place goal-reaching trajectories where CEM can find them. These need not align.
\mtmmse{} and \methodshort{} under-decode block orientation yet still support
substantial planning success, while \accpc{} decodes orientation better but
plans worse. Representation audits and planner behavior should therefore be
reported together: a probe can identify a missing state factor, but it is not a
substitute for closed-loop planning.

\paragraph{Training-only means test-time equivalence.}
The inverse head affects the encoder during optimization, but it is not queried
by CEM and does not propose or score actions at test time. Candidate actions
come entirely from CEM, as in \lewm{}. Thus the empirical differences isolate
training-time representation learning; they do not arise from a stronger
inference procedure or extra runtime computation.

\subsection{Does the model notice impossible transitions?}
\label{sec:surprise}
Goal-reaching success does not by itself show that a latent world model assigns
meaningful surprise to physically inconsistent events. We therefore add a
model-facing diagnostic on PushT and OGBench-Cube. For each sampled trajectory
clip we first record the model's \emph{normal} one-step prediction error: the
latent-space MSE between its prediction and the encoding of the frame that
actually followed. We then corrupt the same clip in one of two ways and measure
the error again: (i) an \emph{action counterfactual} keeps the latent history
but swaps in an action block from a different clip; the model should no longer
expect the observed outcome; and (ii) a \emph{state discontinuity} keeps the
model's prediction but replaces the target future with latents from a different
clip; an impossible jump in the environment state. The reported \emph{surprise
ratio} is the per-clip corrupted error divided by the normal error. A ratio of
$1{\times}$ means the model finds impossible transitions exactly as predictable
as real ones, i.e.\ its dynamics carry no physical content; higher is better.
Both tests use the trained encoder, action encoder, and autoregressive predictor
used by MPC; no decoder or privileged state is introduced.

\begin{table}[t]
\centering
\small
\begin{tabular}{llcc}
\toprule
Task & Model & Action counterfactual ($\uparrow$) & State discontinuity ($\uparrow$) \\
\midrule
PushT & \sigreg{} & $151.7{\pm}4.7{\times}$ & $1246.0{\pm}34.1{\times}$ \\
PushT & \methodshort{} & $9.6{\pm}0.3{\times}$ & $40.0{\pm}1.4{\times}$ \\
OGB-Cube & \sigreg{} & $274.2{\pm}12.6{\times}$ & $835.4{\pm}35.2{\times}$ \\
OGB-Cube & \methodshort{} & $82.3{\pm}1.6{\times}$ & $434.4{\pm}5.6{\times}$ \\
\bottomrule
\end{tabular}
\caption{Latent surprise ratios: one-step prediction error on a corrupted
(physically invalid) transition divided by the same clip's error on the real
transition. $1{\times}$ would mean the model cannot tell invalid transitions
from real ones; higher means the model is more surprised by violations. Each
entry uses 4096 sampled clips per training seed (mean $\pm$ standard deviation
over seeds $\{3072,1,2\}$). In every row, the invalid transition has higher
error than the matched real transition for at least 99.95\% of clips. Ratios
are normalized by each model's own baseline error, so they should be compared
within a model, not across models (see text).}
\label{tab:surprise}
\end{table}

Table~\ref{tab:surprise} shows that both objectives learn action-conditioned
dynamics with strong internal violation signals: every model flags both
corruption types with errors one to three orders of magnitude above its normal
prediction error, and does so on essentially every clip. Two caveats guide the
reading. First, because each ratio is normalized by that model's own
normal-transition error, absolute magnitudes are not comparable across models: a
model with near-zero baseline error can post an enormous ratio without having
better dynamics. The meaningful statements are within-model (\sigreg{} and
\methodshort{} both separate violations from real transitions by large margins)
and qualitative (neither model is fooled). Second, the smaller \methodshort{}
margins on PushT are consistent with its weaker block-state probes and lower
PushT planning success (Section~\ref{sec:mechanisms}), so the diagnostic tracks
the planning evidence rather than contradicting it. These results support a
narrower claim than human-style physical reasoning: the learned latent dynamics
are sensitive to impossible action-conditioned transitions in the same
representation used for planning.

\section{Related Work}

\paragraph{World models and planning.}
The idea of learning a predictive model and planning through it goes back at
least to differentiable recurrent world models for control
\citep{schmidhuber1990worlddifferentiable} and was later popularized in compact
latent form by \citet{ha2018worldmodels}. Reward-driven latent world models such
as Dreamer \citep{hafner2020dreamer} and TD-MPC2 \citep{hansen2024tdmpc2} shape
their representations with reconstruction or value signals; our setting removes
both. We follow the reward-free, goal-conditioned variant: learn from offline
state--action trajectories without task rewards, then plan to a goal image at
test time, which leaves anti-collapse as the central representational problem.

\paragraph{Reconstruction-free latent planning.}
PLDM shows that a JEPA-style latent dynamics model can plan from reward-free
offline data without reconstructing pixels, but relies on a VICReg-derived
multi-term anti-collapse recipe and inverse-dynamics regularization
\citep{sobal2025planning,bardes2022vicreg}. DINO-WM instead uses frozen DINOv2
features and learns a latent dynamics model on top, trading end-to-end
simplicity for strong pretrained visual representations \citep{zhou2025dinowm}.
\lewm{} is the closest predecessor to our work: it trains from pixels end-to-end with a
two-term objective, replacing PLDM's heavier recipe with next-latent prediction
plus \sigreg{} \citep{maes2026leworldmodel,balestriero2025lejepa}. We share
\lewm{}'s simplification agenda but remove its remaining global distributional
assumption during training while leaving its deployed model unchanged. 

SMWM is the most direct prior work: it uses the same basic LeWM encoder,
forward latent predictor, and training-only inverse head, with inverse-action
MSE as the sole anti-collapse term \citep{ivashkov2026smwm}. It evaluates the
same four standard environments and finds that inverse MSE roughly matches
\sigreg{} on the 2D tasks and improves on Cube. We therefore treat \mtmmse{} as
an SMWM-style baseline, not a new mechanism. Our contribution is the
contrastive Action-NCE form, its chance-level collapse bound and Reacher
reliability, a single coefficient held fixed across tasks rather than tuned per
environment, and the Scene and stress-test analysis.

\paragraph{Anti-collapse objectives for JEPAs.}
I-JEPA and V-JEPA stabilize prediction with target encoders, stop-gradient, and
EMA updates \citep{assran2023ijepa,bardes2024vjepa,grill2020byol}. VICReg prevents collapse by
variance and covariance constraints \citep{bardes2022vicreg}. LeJEPA derives a
leaner alternative, \sigreg{}, by regularizing embeddings toward an isotropic
Gaussian through random one-dimensional projections
\citep{balestriero2025lejepa}. These are representation-distribution constraints:
they state what the latent cloud should look like. \mtmmse{} and \methodshort{}
instead state what adjacent latents must \emph{explain}: the action that
connected them. \methodshort{} sharpens this requirement from absolute
regression to contrastive identification. The signal is local,
transition-level, and dynamics-native.

\paragraph{Benchmarks and goal-conditioned baselines.}
The LeWM suite combines top-down navigation, DMC-style reaching, PushT contact
manipulation, and OGBench-Cube \citep{park2025ogbench}. The external baselines
in Figure~\ref{fig:multitask} include PLDM, DINO-WM, goal-conditioned
behavioral cloning, goal-conditioned offline RL, and random policies as reported
by \citet{maes2026leworldmodel}. They place \methodshort{} in the same visual
frame as the original LeWM evaluation; our controlled claims use the released
\lewm{} datasets, architectures, and planner settings so that the comparison
changes the training signal rather than the downstream policy class.

\paragraph{Inverse dynamics and probing.}
Inverse dynamics has long been used to learn controllable representations: as a
self-supervised feature signal for manipulation \citep{agrawal2016poking}, to
build the embedding underlying curiosity rewards \citep{pathak2017curiosity},
as an auxiliary loss for RL \citep{shelhamer2017loss}, and, in multi-step form,
with guarantees that it recovers exactly the control-endogenous part of the
state \citep{lamb2022acro}. PLDM includes an inverse term inside a larger
multi-term objective \citep{sobal2025planning}. Our use is narrower and
sharper: a training-only anti-collapse head attached to an otherwise unchanged
\lewm{} planner, with no privileged state and no planner-time inverse model,
and (unlike ordinary inverse regression) trained to \emph{identify} the
positive action among in-batch continuous-action negatives. Linear probing of
physical quantities is increasingly used to interpret latent world models
\citep{maes2026leworldmodel}; our results caution against inferring planning
quality from such probes alone.

\paragraph{Action-driven contrastive representation learning.}
CLOUD learns forward and inverse dynamics through contrastive estimation for
planning and imitation \citep{wang2021cloud}. \citet{yuan2024actions} use
InfoNCE between visual representations and observed actions to retain
controllable factors, while also using reward prediction and temporal
coherence for online visual RL. TACO instead contrasts current-state and action
sequences against future-state representations for online and offline visual RL
\citep{zheng2023taco}. These works establish action-driven contrast as a useful
representation signal. We use observed continuous actions as the candidate
labels of a training-only inverse task and study that task as the sole
anti-collapse term in an end-to-end, reward-free JEPA world model.

\section{Limitations}
\methodshort{} provides no unconditional geometric guarantee. It requires
variation in the action candidates and useful visual evidence about action
effects; unobserved actuators, stochastic dynamics, no-op-heavy datasets, or
many duplicate actions can weaken the contrastive task. We evaluate normalized
continuous controls only. Discrete, hybrid, structured, and very
high-dimensional action spaces may require different scores, learned action
embeddings, or hard-negative sampling. The loss also introduces training-time
dependence on batch composition, the temperature $\tau$, and one coefficient
$\lambda$; although one setting works across our suite, broader scaling
evidence is needed. PushT shows the flip side of the dynamics-native pressure:
a variable that actions barely move can be underweighted even when it matters
for the task, and exploratory single-seed runs on additional OGBench families
(Appendix~\ref{sec:appendix-broader-ogb}) suggest the same trade-off on
combinatorial puzzle tasks.

\methodshort{} improves several goal-conditioned planning protocols and shows
strong latent surprise on counterfactual-action and state-discontinuity probes,
but these diagnostics remain internal to the trained latent dynamics. They do
not replace public violation-of-expectation suites with human-designed physical
events or prove robust out-of-distribution physical reasoning. OGBench Visual
Scene is also evaluated under our trajectory-goal protocol rather than the
official five-task fixed-goal OGBench leaderboard protocol, so its 80\% success
rate should be read as a matched stress test, not a public benchmark score;
under the official 750-step fixed-goal protocol, neither method solves the
tasks at this model scale (Appendix~\ref{sec:appendix-scene-audit}).

\section{Conclusion}
A world model's representation has to be protected from collapse, but nothing
about the problem says that protection must take the form of a prescribed
latent distribution. \methodshort{} replaces \lewm{}'s Gaussian matching with a
contrastive, dynamics-native signal: keep whatever information identifies the
action that drove each transition. The result is still a one-stage, end-to-end
pixel JEPA without target network, stop-gradient branch, frozen encoder,
reconstruction decoder, or fixed global latent distribution, and the deployed
model is unchanged. The empirical picture is not universal dominance: PushT
exposes a real limitation of inverse-action signals when task-relevant state is
weakly controlled. Where the environment couples many controllable factors,
our results are consistent with prescribed geometry becoming a bottleneck: on
OGBench Visual Scene, \sigreg{} reaches 58.0\% while \methodshort{} reaches
80.0\% across matched seeds, though the Scene gain may also reflect lower
one-step forward error. On
TwoRoom-long \methodshort{} retains most of the inverse-dynamics gain over
\sigreg{}. These results show that transition-derived anti-collapse can replace
a fixed global latent prior while preserving planning quality.

\bibliographystyle{iclr2026_conference}
\bibliography{custom}

@article{lecun2022path,
  title={A Path Towards Autonomous Machine Intelligence},
  author={LeCun, Yann},
  journal={OpenReview},
  year={2022}
}

@inproceedings{bardes2022vicreg,
  title={{VICReg}: Variance-Invariance-Covariance Regularization for Self-Supervised Learning},
  author={Bardes, Adrien and Ponce, Jean and LeCun, Yann},
  booktitle={International Conference on Learning Representations (ICLR)},
  year={2022}
}

@inproceedings{assran2023ijepa,
  title={Self-Supervised Learning from Images with a Joint-Embedding Predictive Architecture},
  author={Assran, Mahmoud and Duval, Quentin and Misra, Ishan and Bojanowski, Piotr and Vincent, Pascal and Rabbat, Michael and LeCun, Yann and Ballas, Nicolas},
  booktitle={IEEE/CVF Conference on Computer Vision and Pattern Recognition (CVPR)},
  year={2023}
}

@article{bardes2024vjepa,
  title={Revisiting Feature Prediction for Learning Visual Representations from Video},
  author={Bardes, Adrien and Garrido, Quentin and Ponce, Jean and Chen, Xinlei and Rabbat, Michael and LeCun, Yann and Assran, Mahmoud and Ballas, Nicolas},
  journal={Transactions on Machine Learning Research},
  year={2024}
}

@inproceedings{zhou2025dinowm,
  title={{DINO-WM}: World Models on Pre-trained Visual Features enable Zero-shot Planning},
  author={Zhou, Gaoyue and Pan, Hengkai and LeCun, Yann and Pinto, Lerrel},
  booktitle={Proceedings of the 42nd International Conference on Machine Learning},
  series={Proceedings of Machine Learning Research},
  volume={267},
  pages={79115--79135},
  year={2025}
}

@inproceedings{sobal2025planning,
  title={Learning from Reward-Free Offline Data: A Case for Planning with Latent Dynamics Models},
  author={Sobal, Uladzislau and Zhang, Wancong and Cho, Kyunghyun and Balestriero, Randall and Rudner, Tim G. J. and LeCun, Yann},
  booktitle={Advances in Neural Information Processing Systems},
  volume={38},
  year={2025}
}

@article{balestriero2025lejepa,
  title={{LeJEPA}: Provable and Scalable Self-Supervised Learning Without the Heuristics},
  author={Balestriero, Randall and LeCun, Yann},
  journal={arXiv preprint arXiv:2511.08544},
  year={2025}
}

@article{maes2026leworldmodel,
  title={{LeWorldModel}: Stable End-to-End Joint-Embedding Predictive Architecture from Pixels},
  author={Maes, Lucas and Le Lidec, Quentin and Scieur, Damien and LeCun, Yann and Balestriero, Randall},
  journal={arXiv preprint arXiv:2603.19312},
  year={2026}
}

@article{ivashkov2026smwm,
  title={Sensorimotor World Models: Perception for Action via Inverse Dynamics},
  author={Ivashkov, Petr and Balestriero, Randall and Sch{\"o}lkopf, Bernhard},
  journal={arXiv preprint arXiv:2606.20104},
  year={2026}
}

@techreport{schmidhuber1990worlddifferentiable,
  title={Making the World Differentiable: On Using Self-Supervised Fully Recurrent Neural Networks for Dynamic Reinforcement Learning and Planning in Non-Stationary Environments},
  author={Schmidhuber, J{\"u}rgen},
  institution={Institut f{\"u}r Informatik, Technische Universit{\"a}t M{\"u}nchen},
  number={FKI-126-90},
  year={1990}
}

@inproceedings{ha2018worldmodels,
  title={Recurrent World Models Facilitate Policy Evolution},
  author={Ha, David and Schmidhuber, J{\"u}rgen},
  booktitle={Advances in Neural Information Processing Systems},
  year={2018}
}

@inproceedings{park2025ogbench,
  title={{OGBench}: Benchmarking Offline Goal-Conditioned Reinforcement Learning},
  author={Park, Seohong and Frans, Kevin and Eysenbach, Benjamin and Levine, Sergey},
  booktitle={International Conference on Learning Representations},
  year={2025}
}

@inproceedings{gutmann2010nce,
  title={Noise-Contrastive Estimation: A New Estimation Principle for Unnormalized Statistical Models},
  author={Gutmann, Michael and Hyv{\"a}rinen, Aapo},
  booktitle={International Conference on Artificial Intelligence and Statistics (AISTATS)},
  year={2010}
}

@article{oord2018cpc,
  title={Representation Learning with Contrastive Predictive Coding},
  author={van den Oord, A{\"a}ron and Li, Yazhe and Vinyals, Oriol},
  journal={arXiv preprint arXiv:1807.03748},
  year={2018}
}

@inproceedings{agrawal2016poking,
  title={Learning to Poke by Poking: Experiential Learning of Intuitive Physics},
  author={Agrawal, Pulkit and Nair, Ashvin and Abbeel, Pieter and Malik, Jitendra and Levine, Sergey},
  booktitle={Advances in Neural Information Processing Systems},
  year={2016}
}

@inproceedings{pathak2017curiosity,
  title={Curiosity-Driven Exploration by Self-Supervised Prediction},
  author={Pathak, Deepak and Agrawal, Pulkit and Efros, Alexei A. and Darrell, Trevor},
  booktitle={International Conference on Machine Learning (ICML)},
  year={2017}
}

@article{shelhamer2017loss,
  title={Loss is its own Reward: Self-Supervision for Reinforcement Learning},
  author={Shelhamer, Evan and Mahmoudieh, Parsa and Argus, Max and Darrell, Trevor},
  journal={arXiv preprint arXiv:1612.07307},
  year={2016}
}

@article{lamb2022acro,
  title={Guaranteed Discovery of Control-Endogenous Latent States with Multi-Step Inverse Models},
  author={Lamb, Alex and Islam, Riashat and Efroni, Yonathan and Didolkar, Aniket and Misra, Dipendra and Foster, Dylan and Molu, Lekan and Chari, Rajan and Krishnamurthy, Akshay and Langford, John},
  journal={Transactions on Machine Learning Research},
  year={2023}
}

@inproceedings{hafner2020dreamer,
  title={Dream to Control: Learning Behaviors by Latent Imagination},
  author={Hafner, Danijar and Lillicrap, Timothy and Ba, Jimmy and Norouzi, Mohammad},
  booktitle={International Conference on Learning Representations (ICLR)},
  year={2020}
}

@inproceedings{hansen2024tdmpc2,
  title={{TD-MPC2}: Scalable, Robust World Models for Continuous Control},
  author={Hansen, Nicklas and Su, Hao and Wang, Xiaolong},
  booktitle={International Conference on Learning Representations (ICLR)},
  year={2024}
}

@inproceedings{grill2020byol,
  title={Bootstrap Your Own Latent: A New Approach to Self-Supervised Learning},
  author={Grill, Jean-Bastien and Strub, Florian and Altch{\'e}, Florent and Tallec, Corentin and Richemond, Pierre H. and Buchatskaya, Elena and Doersch, Carl and Pires, Bernardo Avila and Guo, Zhaohan Daniel and Azar, Mohammad Gheshlaghi and Piot, Bilal and Kavukcuoglu, Koray and Munos, R{\'e}mi and Valko, Michal},
  booktitle={Advances in Neural Information Processing Systems},
  year={2020}
}

@inproceedings{wang2021cloud,
  title={{CLOUD}: Contrastive Learning of Unsupervised Dynamics},
  author={Wang, Jianren and Lu, Yujie and Zhao, Hang},
  booktitle={Proceedings of the 2020 Conference on Robot Learning},
  series={Proceedings of Machine Learning Research},
  volume={155},
  pages={365--376},
  year={2021}
}

@article{yuan2024actions,
  title={Learning Task-Relevant Representations via Rewards and Real Actions for Reinforcement Learning},
  author={Yuan, Linghui and Lu, Xiaowei and Liu, Yunlong},
  journal={Knowledge-Based Systems},
  volume={294},
  pages={111788},
  doi={10.1016/j.knosys.2024.111788},
  year={2024}
}

@inproceedings{zheng2023taco,
  title={{TACO}: Temporal Latent Action-Driven Contrastive Loss for Visual Reinforcement Learning},
  author={Zheng, Ruijie and Wang, Xiyao and Sun, Yanchao and Ma, Shuang and Zhao, Jieyu and Xu, Huazhe and Daum{\'e} III, Hal and Huang, Furong},
  booktitle={Advances in Neural Information Processing Systems},
  volume={36},
  year={2023}
}

\appendix

\section{Evaluation Protocol}
\label{sec:appendix-protocol}
Our controlled standard-task planning tables use 200 episodes, evaluation seed
42, exact checkpoint stems, CEM 300 samples / 30 elites / 30 iterations, and
unique output files. \sigreg{}, \mtmmse{}, and \methodshort{} use final epoch-10
checkpoints from training seeds $\{3072,1,2\}$. No validation-based checkpoint
selection is used in the controlled tables. The
non-\lewm{} baselines in Figure~\ref{fig:multitask} are paper-reported values from
\citet{maes2026leworldmodel}; they are included for continuity with the original
evaluation style, not as reruns.

\begin{table}[ht]
\centering
\small
\begin{tabularx}{\textwidth}{lX}
\toprule
Task & Controlled evaluation decision \\
\midrule
TwoRoom & Use released/maintainer-confirmed goal offset $25$, budget $50$.
Issue \#38 records maintainer confirmation that the paper's $100/150$ text was a
typo; issue \#67 documents the same mismatch and low success under $100/150$.
We keep $100/150$ only as the separately labeled TwoRoom-long stress test. \\
PushT & Use released config: goal offset $25$, budget $50$, CEM $300/30/30$. \\
Reacher & Use released config: goal offset $25$, budget $50$, CEM $300/30/30$.
Issue \#41 records that the paper text says $10$ CEM iterations for non-PushT
tasks while the released config uses $30$; no maintainer correction is visible.
Issues \#37 and \#62 show Reacher sensitivity to training/eval seed. \\
OGBench-Cube & Use released config: goal offset $25$, budget $50$, CEM
$300/30/30$, with the same CEM caveat as Reacher from issue \#41. \\
OGBench-Scene & Added complex-task trajectory-goal protocol using
\texttt{visual-scene-play-v0}: goal offset $25$, budget $50$, CEM $300/30/30$, 50
episodes per training seed, and matched train seeds $\{3072,1,2\}$ for both
\sigreg{} and \methodshort{}. This is separate from the official OGBench
Visual Scene five-task fixed-goal protocol. \\
\bottomrule
\end{tabularx}
\caption{Protocol audit for the 200-episode controlled comparisons.
Issue numbers refer to the public \lewm{} repository:
\url{https://github.com/lucas-maes/le-wm/issues}:
\#38, \#67, \#41, \#37, and \#62.}
\label{tab:protocol-audit}
\end{table}

\section{Interrogating the OGBench Scene Result}
\label{sec:appendix-scene-audit}
The Scene margin is large enough that it deserves a separate audit. Table
\ref{tab:scene-audit} lists the checks we performed on the completed
trajectory-goal runs. The headline survives these checks, but the causal
interpretation should remain narrow: this is evidence about our matched
trajectory-goal MPC protocol, not the public OGBench fixed-goal benchmark.

\begin{table}[ht]
\centering
\small
\begin{tabularx}{\textwidth}{lXX}
\toprule
Check & Finding & Interpretation \\
\midrule
Episode pairing & All six result files use evaluation seed 42, 50 episodes,
goal offset 25, budget 50, CEM $300/30/30$, horizon 5, and action block 5. The
raw success arrays give 40 \methodshort{}-only wins and 7 \sigreg{}-only wins
over 150 paired episodes. & The gap is not an artifact of comparing different
episode sets. \\
Training comparability & The resolved training configs match on dataset
\texttt{ogbench/visual\_scene\_play}, batch size 128, learning rate
$5{\times}10^{-5}$, 10 epochs, history size 3, encoder, action encoder, and
forward predictor. They differ in the intended objective: \sigreg{} weight 0.09
versus \actionnce{} with inverse weight 0.30 and no \sigreg{}. & The comparison
isolates the anti-collapse/training signal up to the small training-only inverse
head. \\
Checkpoint selection & Eval logs load the epoch-10 object checkpoint for both
\sigreg{} and \methodshort{}. & The result is not caused by accidentally
evaluating an earlier or best-selected checkpoint. \\
Privileged-state leakage & The Scene evaluator uses privileged future state to
reset simulator targets and define success, but the world-model cost path encodes
goal pixels and drops goal action before computing final latent distance. & This
matches the trajectory-goal setup: privileged state defines the task, not the
planner input. \\
Collapse & Final training logs show nonzero latent scale for both methods
(roughly 0.99 mean embedding standard deviation for \sigreg{} and 0.24 for
\methodshort{} over the last logged batches). & The \sigreg{} model is not
constant-collapsed; the gap is a planner-usable-geometry or dynamics-quality
failure, not trivial collapse. \\
Protocol scope & The goal is a future visual observation from
\texttt{visual-scene-play-v0}; public OGBench Scene tables average five fixed
goal tasks with a 750-step cap. & The 80\% number should not be compared to
public OGBench leaderboard scores. \\
\bottomrule
\end{tabularx}
\caption{Audit of the completed OGBench Visual Scene
trajectory-goal comparison.}
\label{tab:scene-audit}
\end{table}

\paragraph{Why can the gap be this large?}
Scene couples several slow variables (drawer, window, two buttons, and a cube
state) to the same arm observation. A final-latent MPC cost is brittle to
dropping any one of these variables: a rollout can be close in agent pose while
still missing the target object configuration. \sigreg{} prevents constant
collapse by enforcing a global isotropic latent distribution, but that guarantee
does not say which state factors are preserved with a geometry useful for final
latent distance. In a multi-object scene, part of the latent budget can be spent
matching the global marginal or representing visually variable but
planning-irrelevant factors.

\methodshort{} supplies a different pressure. To identify the observed action
among in-batch alternatives, adjacent latents must retain information about
controllable changes. This favors variables whose changes explain the executed
control and therefore aligns more directly with the CEM rollout interface. The
training logs also show that \methodshort{} reaches a lower final one-step
forward MSE on Scene (about $0.002$ versus $0.004$ for \sigreg{} in the last
logged batches), so the observed advantage may combine better transition
geometry with more accurate short-horizon latent dynamics. Consistent with this
account, the non-contrastive \mtmmse{} ablation trained under the identical
Scene protocol also clears \sigreg{} (75.3\% vs.\ 58.0\%, paired 39/13,
$p\!\approx\!4{\times}10^{-4}$), so the win belongs to inverse-dynamics
supervision as a family, with the contrastive form contributing a further
(noise-level) 4.7 points. The result should therefore be interpreted as a
downstream planning win for dynamics-derived training signals, not as a pure
proof that latent marginal normality alone is harmful.

\paragraph{The official fixed-goal protocol.}
For completeness we also ran the official OGBench Visual Scene protocol (five
fixed goal tasks, 750-step cap) with the matched seed-3072 checkpoints: both
\sigreg{} and \methodshort{} score $0/250$, with every episode reaching the
step cap. The official tasks demand much longer-horizon manipulation than the
$25$-step trajectory goals used in training-matched evaluation, and the
final-latent CEM planner does not solve them at this model scale for either
objective. This is why Table~\ref{tab:scene} reports the matched
trajectory-goal protocol: it is the regime in which these world models operate,
and it keeps the comparison between training signals rather than between
planners and task horizons.

\paragraph{What would falsify the interpretation?}
Two further checks would make the Scene story tighter. First, an
evaluation-seed sweep or larger $n$ would test whether the same paired
advantage holds beyond the current 150 matched episodes. Second, per-factor
diagnostics (drawer, window, button, and cube target probes or per-target
success decomposition) would show which Scene variables \sigreg{} loses.

Probe details: $n{=}4000$ frames, ridge $\alpha{=}1$, a $70/30$ split, and
per-dimension $\rtwo{}$ on privileged state. PushT probes agent $x,y$, block
$x,y$, block orientation, and two velocity components. All probe rows shown use
epoch-10 checkpoints.

\paragraph{CEM action-block semantics.}
At test time, no observed action is supplied. The planner samples actions in the
same normalized action space used by the world model. With raw action dimension
$d$, planning horizon $H{=}5$, and action block $K{=}5$, CEM samples candidate
sequences $A\in\mathbb{R}^{B\times 300\times H\times Kd}$. Each of the $H$
coarse planner slots is therefore a flattened block of $K$ low-level simulator
actions. CEM initializes a diagonal Gaussian over these blocks, samples 300
candidate sequences, inserts the current mean as one candidate, rolls the world
model forward under every candidate, keeps the 30 lowest final latent-goal MSE
sequences, and refits the Gaussian mean and per-component scale to those elites.
This sample-score-refit loop is repeated for 30 iterations, after which the
policy returns the final mean sequence.

\begin{figure}[t]
\centering
\includegraphics[width=\linewidth]{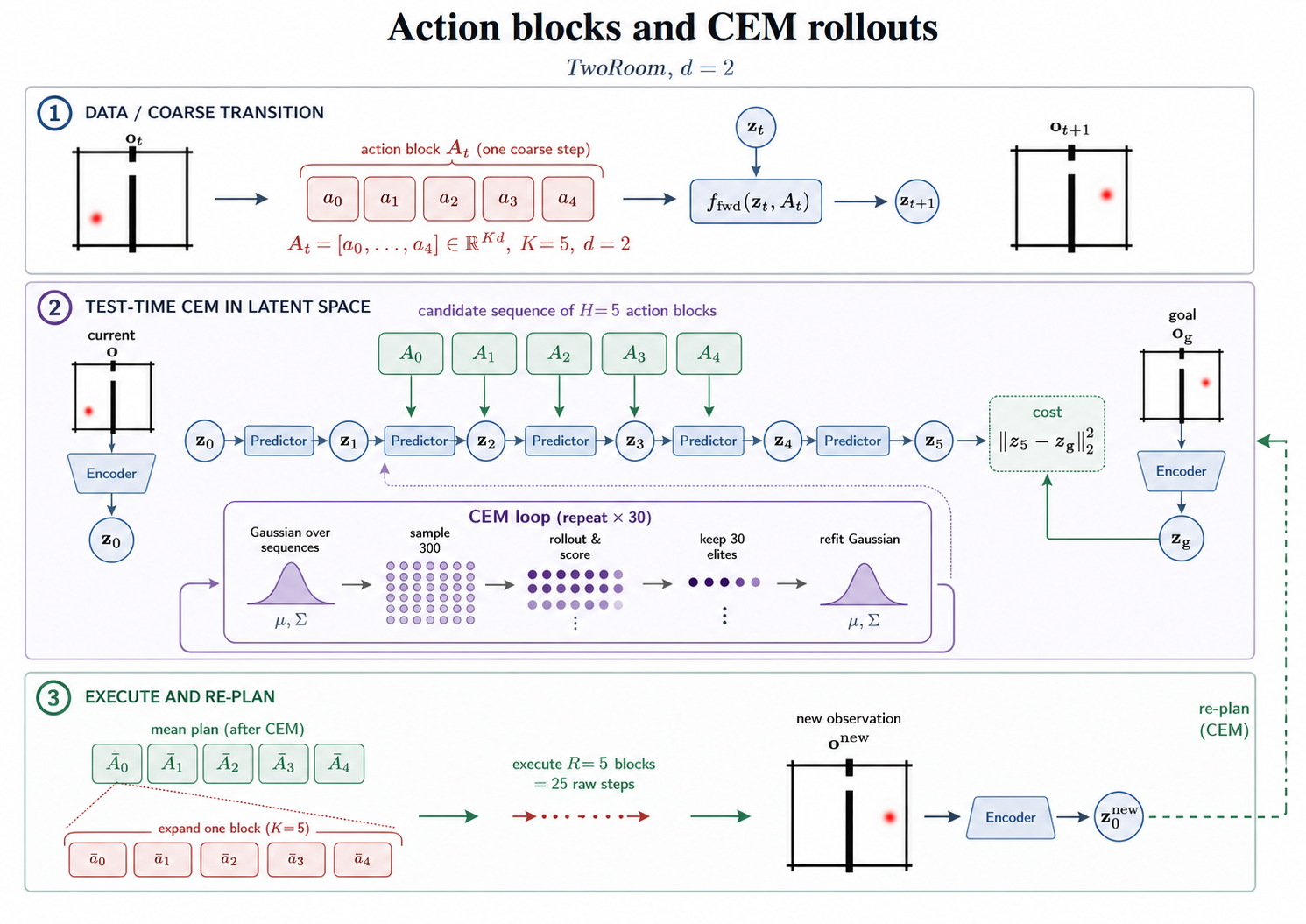}
\caption{Action blocks and CEM rollouts (TwoRoom, $d{=}2$). Each coarse
transition bundles $K{=}5$ raw simulator actions into one model input block.
CEM samples $H{=}5$-block candidate sequences, rolls them through the unchanged
latent forward model, scores final latent-goal distance, refits to elites, and
executes the final mean blocks as ordered low-level \texttt{env.step} actions
before re-planning.}
\label{fig:action-block}
\end{figure}

Figure~\ref{fig:action-block} summarizes how the three time scales fit together.
They coincide at the value five in this configuration and should not be
conflated. The \emph{action block} $K{=}5$ is the number of raw
simulator steps bundled into one coarse transition (equal to the dataset
frameskip); the \emph{planning horizon} $H{=}5$ is the number of coarse
transitions the planner looks ahead, i.e.\ $HK{=}25$ simulator steps; and the
\emph{receding horizon} $R{=}5$ is the number of coarse transitions executed
before the policy re-plans, i.e.\ $RK{=}25$ environment steps per replan. Here
$R{=}H$, so each plan is executed in full before re-planning; smaller $R$ yields
tighter closed-loop control and warm-starts the next solve from the unused tail
of the current plan. The world model predicts the next coarse \emph{latent
state}, not actions and not the $K{-}1$ skipped frames: one forward step maps the
latent of coarse frame $t$ to that of coarse frame $t{+}1$ conditioned on the
entire $K$-action block, and only the final predicted latent is scored against
the goal. The action block is thus the optimization variable supplied by CEM at
test time and by the dataset at training time. It is never an output of the model.

For example, on a two-dimensional action task with $K{=}5$, one coarse action
slot is
\[
[a^x_0,a^y_0,a^x_1,a^y_1,a^x_2,a^y_2,a^x_3,a^y_3,a^x_4,a^y_4].
\]
This slot connects adjacent model frames separated by five simulator steps. The
policy reshapes it into the ordered low-level sequence
$(a_0,a_1,a_2,a_3,a_4)$, inverse-normalizes those five actions, and executes
them with five consecutive environment steps. Thus the action associated with a
coarse transition is not one command repeated for five frames; it is the whole
intervening action block. The HDF5 training loader uses the same convention:
pixels and other non-action columns are subsampled every $K$ steps, while the
action column is kept dense and reshaped into one flattened block per coarse
transition.

\paragraph{Cold start and self-conditioning.}
The rollout never receives an empty context. On episode reset the frame-history
buffer is pre-filled by repeating the initial observation, so the policy always
supplies a full \texttt{history\_size}-length context; for TwoRoom evaluation
\texttt{history\_size}${=}1$, that context is the single current frame. The
predictor is a Transformer, so a length-one sequence is valid input: it encodes the
initial frame to $\z_0$ and predicts $\z_1$ from $\z_0$ and the first action block.
Thereafter the autoregressive rollout conditions on its \emph{own} predicted
latents until the window reaches the trained length and then slides, so beyond the
first step the history is imagined rather than observed, which is precisely why a
stable latent geometry matters, as an early error propagates through the remaining
coarse steps. The autoregressive model trains on three real context frames but
TwoRoom evaluation begins from one, a mild context-length difference the
variable-length predictor tolerates.

\paragraph{CEM distribution updates.}
The sampling distribution is a \emph{diagonal} Gaussian over the flattened plan:
for TwoRoom the plan is $H{\times}Kd=5{\times}10$, so the mean and scale are each
$5{\times}10$ tensors; fifty independent one-dimensional Gaussians per
environment, with no cross-coordinate covariance. Each iteration draws $300$ plans
$A_s=\mu+\sigma\odot\varepsilon_s$ with $\varepsilon_s\sim\mathcal{N}(0,I)$, forces
$A_0{=}\mu$, scores each by final-latent goal distance, and refits $\mu$ and
$\sigma$ to the per-coordinate mean and standard deviation of the $30$ elite plans.
Selection is \emph{joint} (whole plans are ranked) but the refit is \emph{marginal}
(each coordinate's new moments are those of the elites at that coordinate), so a
coordinate that drives the cost is sharpened while an irrelevant one stays diffuse.
For instance, a coordinate whose useful value is ${\approx}{+}0.7$ might move
$\mu:0.00\!\to\!0.42\!\to\!0.61\!\to\!0.69$ with $\sigma$ contracting
$1.00\!\to\!0.55\!\to\!0.30\!\to\!0.16$ over the first iterations and freezing near
$\mu{\approx}0.71$, $\sigma{\approx}0.04$, whereas an unleveraged coordinate keeps
$\mu{\approx}0$, $\sigma{\approx}1$. The refit is \emph{hard}. Moments are
replaced, with no step size or variance floor, so the Gaussian collapses to nearly
a point within roughly fifteen iterations.

\paragraph{Training-only action contrast.}
For each $B\times T$ batch, the inverse task flattens
$N=B(T-1)$ adjacent transitions and uses all observed actions in that flattened
batch as candidates in Equation~\ref{eq:action-nce}. Each query
$\hat{\aact}_i=\inv(\z_i,\z_{i+1})$ and candidate $\aact_j$ is a full coarse action
block of dimension $Kd$ (all $K$ raw actions, e.g.\ $10$ on TwoRoom), regressed in
a single pass from the latent pair rather than decoded into $K$ ordered actions;
the horizon-conditioned variant instead predicts the first block $\aact_t$ from a
$k$-step-apart pair $(\z_t,\z_{t+k},e_k)$. The inverse head is
optimized and checkpointed for reproducibility, but rollout and cost functions
never call it; removing it after training leaves predictions and CEM action
selection unchanged.

\section{Controlled Three-Way Ablation}
\label{sec:appendix-controlled}

\begin{figure}[t]
\centering
\includegraphics[width=0.92\linewidth]{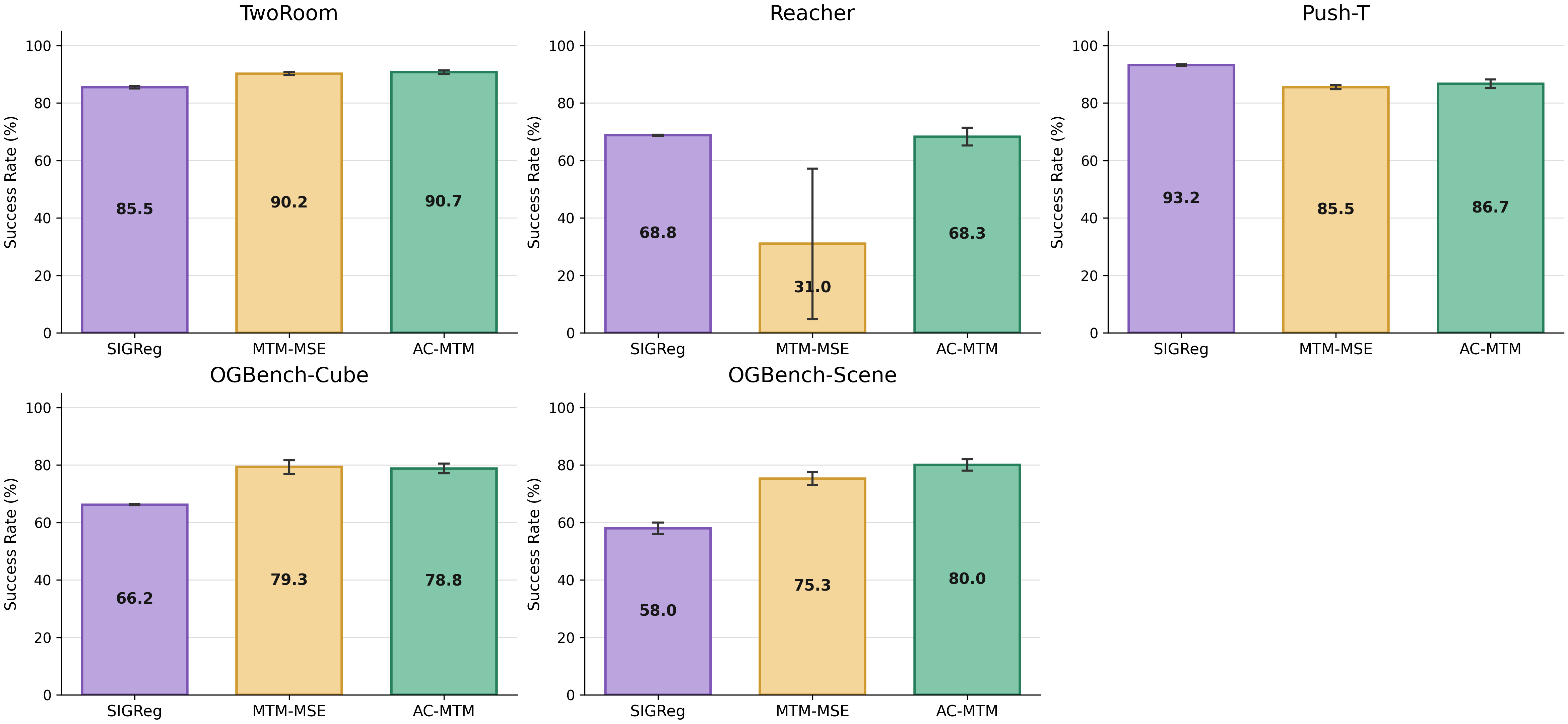}
\caption{Controlled three-way ablation with 200 evaluation episodes. Error bars
show standard deviation over three training seeds. The Reacher panel shows why
\actionnce{} is used as the main method: it removes the bimodal collapse of the
non-contrastive inverse-regression ablation without changing the test-time
planner.}
\label{fig:ablation-threeway}
\end{figure}

Table~\ref{tab:controlled-threeway} and Figure~\ref{fig:ablation-threeway} give
the matched comparison between \sigreg{}, the non-contrastive
inverse-regression ablation, and \methodshort{}. This diagnostic isolates the
effect of \actionnce{}: on TwoRoom, PushT, and Cube, \methodshort{} remains
within 1.2 percentage points of \mtmmse{}; on Reacher, \mtmmse{} collapses for
two of three seeds whereas \methodshort{} remains non-collapsed; on Scene,
both inverse variants beat \sigreg{} decisively and \methodshort{} leads
\mtmmse{} by 4.7 points (paired 16/9, $p\!\approx\!0.23$).

\begin{table}[ht]
\centering
\small
\begin{tabular}{lccc}
\toprule
Task & \sigreg{} (\lewm{}) & \mtmmse{} & \methodshort{} (ours) \\
\midrule
TwoRoom  & $85.5{\pm}0.4$ & $90.2{\pm}0.5$ & $\mathbf{90.7{\pm}0.6}$ \\
Reacher  & $\mathbf{68.8{\pm}0.2}$ & $31.0{\pm}26.2$ & $68.3{\pm}3.1$ \\
PushT    & $\mathbf{93.2{\pm}0.2}$ & $85.5{\pm}0.7$ & $86.7{\pm}1.5$ \\
OGB-Cube & $66.2{\pm}0.2$ & $\mathbf{79.3{\pm}2.4}$ & $78.8{\pm}1.7$ \\
OGB-Scene & $58.0{\pm}2.0$ & $75.3{\pm}2.3$ & $\mathbf{80.0{\pm}2.0}$ \\
\bottomrule
\end{tabular}
\caption{Controlled planning success (\%) with the shared autoregressive
planner: 200 evaluation episodes per seed on the standard tasks and 50 on
OGBench-Scene. Values are mean $\pm$ standard deviation over three training
seeds.}
\label{tab:controlled-threeway}
\end{table}

\section{Exploratory Single-Seed Results on Additional OGBench Families}
\label{sec:appendix-broader-ogb}

To probe how the two training signals behave outside the main suite, we ran
single-seed (3072) comparisons on six additional OGBench environments under the
same trajectory-goal protocol (50 episodes, evaluation seed 42). These runs use
one training seed and no tuning, so they are reported as exploratory scope
evidence, not controlled claims.

\begin{table}[ht]
\centering
\small
\begin{tabular}{lccc}
\toprule
Task & \sigreg{} & \methodshort{} & $\Delta$ \\
\midrule
Visual Puzzle 4x4 (play) & \textbf{50.0} & 34.0 & \textcolor{loss}{$-16$} \\
Visual Puzzle 4x5 (play) & \textbf{52.0} & 26.0 & \textcolor{loss}{$-26$} \\
Visual AntMaze teleport (navigate) & 40.0 & \textbf{46.0} & \textcolor{win}{$+6$} \\
Powderworld medium (play) & 6.0 & \textbf{16.0} & \textcolor{win}{$+10$} \\
Visual AntMaze large (stitch) & 30.0 & 30.0 & $0$ \\
AntSoccer medium (stitch) & 88.0 & 88.0 & $0$ \\
\bottomrule
\end{tabular}
\caption{Exploratory single-seed (3072) trajectory-goal success (\%) on
additional OGBench families, 50 episodes each. \sigreg{} is clearly better on
the combinatorial button-puzzle tasks, \methodshort{} modestly better on the
stochastic-teleport maze and Powderworld, and the two tie on the stitching
tasks.}
\label{tab:broader-ogb}
\end{table}

The pattern is consistent with the PushT analysis in
Section~\ref{sec:mechanisms}. The puzzle tasks are dominated by a grid of
buttons whose visual state changes discretely and near-identically regardless
of which action toggled them; identifying the executed action then provides
little pressure to represent the full button configuration, whereas \sigreg{}'s
global variance pressure preserves it. Frozen-latent probes confirm this
mechanism: on Puzzle 4x4, \sigreg{} decodes the variable button bits at
$\rtwo{}\!\approx\!0.98$ while \methodshort{} decodes them at chance
($\rtwo{}\!\le\!0$), even though both models decode the continuous arm pose at
$\rtwo{}\!\approx\!0.99$. Conversely, where the environment adds
stochastic or diffuse dynamics (teleporting maze, falling-particle
Powderworld), the dynamics-native signal is at least as good. We report these
runs to delimit the method's scope: the claim is not that contrastive inverse
dynamics dominates distributional regularization everywhere, but that it is a
robust distribution-free alternative whose advantage grows with the number of
coupled controllable factors, as in Scene.

\end{document}